\pdfoutput=1
\documentclass[11pt,a4paper]{article}
\usepackage[hyperref]{acl2021}
\usepackage{times}
\usepackage{latexsym}

\usepackage{longtable}
\usepackage{comment}
\usepackage{nicefrac}
\usepackage{amsmath}
\usepackage{amssymb}
\usepackage{amsfonts}
\usepackage{amsthm}
\usepackage{mathtools}
\usepackage{booktabs}
\usepackage{tabularx}
\usepackage{multirow}
\usepackage{relsize}
\usepackage{longtable}
\usepackage{xtab}
\usepackage{xspace}
\usepackage{enumitem}
\usepackage{textcomp}
\usepackage{chemarrow}
\usepackage{graphicx}
\usepackage{stmaryrd}
\usepackage{pifont}
\usepackage{colortbl}
\usepackage{tabstackengine}
\usepackage{adjustbox}
\usepackage{tikz-dependency}
\usepackage{stackengine, scalerel}
\usepackage{calc}
\usepackage{float}
\usepackage{thmtools}

\usepackage{microtype}

\aclfinalcopy 

\usepackage{subcaption}

\usepackage{tikz}
\usetikzlibrary{arrows.meta}
\usetikzlibrary{shapes.arrows}
\usetikzlibrary{shapes.misc}
\usetikzlibrary{fit}

\usepackage{algorithm}
\usepackage{algorithmicx}
\usepackage[noend]{algpseudocode}
\algnewcommand{\parState}[1]{\State%
    \parbox[t]{\dimexpr\linewidth-\algmargin}{\strut\hangindent=\algorithmicindent \hangafter=1 #1\strut}}
\algrenewcommand\algorithmicindent{1.0em}%
\newcommand{\algorithmicdowhile}{\textbf{do}:}
\algdef{SE}[DOWHILE]{Do}{doWhile}{\algorithmicdowhile}[1]{\algorithmicwhile\ #1}%
\newcommand{\algorithmicfunc}[1]{\textbf{def} #1 :}
\algdef{SE}[FUNC]{Func}{EndFunc}[1]{\algorithmicfunc{#1}}{}

\newif\ifboldnumber

\algrenewcommand\alglinenumber[1]{%
  \footnotesize\ifboldnumber\color{red}\bfseries\fi\global\boldnumberfalse#1:}

\MakeRobust{\Call}   

\newcommand{\rightcomment}[1]{{\color{commentcolor} \(\triangleright\) {\footnotesize\textit{#1}}}}
\algrenewcommand{\algorithmiccomment}[1]{\hfill \rightcomment{#1}}  
\algnewcommand{\LineComment}[1]{\State \rightcomment{#1}}
\algnewcommand{\LinesComment}[1]{\State \rightcomment{\parbox[t]{\linewidth-\leftmargin-\widthof{\(\triangleright\) }}{#1}}}

\renewcommand\algorithmicthen{:}

\makeatletter
\ifthenelse{\equal{\ALG@noend}{t}}%
  {\algtext*{EndFunc}}
  {}%
\makeatother

\algnewcommand{\IIf}[1]{\State\algorithmicif\ #1\ \algorithmicthen}
\algnewcommand{\EndIIf}{\unskip}

\usepackage{emoji}

\let\rightarrow\chemarrow 

\newtheorem{cor}{Corollary}

\usepackage{thmtools}
\usepackage{thm-restate}
\usepackage{apxproof}

\theoremstyle{definition}
\newtheorem{definition}{Definition}

\usepackage{cleveref}
\crefname{section}{\S}{\S\S}
\Crefname{section}{\S}{\S\S}
\crefname{table}{Tab.}{}
\crefname{figure}{Fig.}{}
\crefname{algorithm}{Alg}{}
\crefname{algorithm}{Alg}{}
\crefname{line}{Line}{}
\crefname{appendix}{App.}{}
\crefformat{section}{\S#2#1#3}
\crefname{thm}{Theorem}{}
\crefname{prop}{Proposition}{}
\crefname{defin}{Definition}{}
\crefname{lemma}{Lemma}{}
\crefname{cor}{Corollary}{}
\crefname{equation}{}{}

\makeatletter
\g@addto@macro\normalsize{%
  \setlength\abovedisplayskip{5pt}
  \setlength\belowdisplayskip{5pt}
  \setlength\abovedisplayshortskip{5pt}
  \setlength\belowdisplayshortskip{5pt}
}
\makeatother

\renewcommand{\paragraph}[1]{\noindent\textbf{#1}\quad}

\definecolor{forestgreen}{rgb}{0.13, 0.55, 0.13}
\newcommand{\Cma}{forestgreen}
\definecolor{verydarkgray}{gray}{0.45}
\newcommand{\Cdead}{verydarkgray}
\definecolor{mediumblue}{rgb}{0.0, 0.0, 0.8}
\newcommand{\Cselect}{mediumblue}
\definecolor{darkgray}{gray}{0.6}
\newcommand{\Cexc}{darkgray}
\definecolor{crimson}{rgb}{0.86, 0.08, 0.24}
\definecolor{commentcolor}{rgb}{0.4, 0.22, 0.33}

\newcommand{\checkNotation}[1]{{\color{blue}#1}}
\renewcommand{\checkNotation}[1]{{#1}}

\newcommand{\defeq}[0]{\overset{\smaller\mathrm{def}}{=}}
\newcommand{\defn}[1]{\textbf{#1}}
\renewcommand{\th}{^{\text{th}}}
\renewcommand{\setminus}{\smallsetminus}
\renewcommand{\bar}[1]{\overline{#1}}

\newcommand{\bigo}[1]{\checkNotation{\mathcal{O}(#1)}}

\newcommand{\msetminus}[0]{{\backslash\!\!\backslash}}

\newcommand{\medgedelete}[2]{\checkNotation{#1 \msetminus #2}}
\newcommand{\case}[1]{\checkNotation{\noindent\emph{#1:}}}

\newcommand{\nN}{\checkNotation{N}}
\newcommand{\nK}{\checkNotation{K}}

\DeclareMathOperator*{\argmax}{argmax\,}
\DeclareMathOperator*{\argmin}{argmin\,}
\newcommand{\abs}[1]{\lvert #1 \rvert}

\newcommand{\graph}{\checkNotation{G}}
\newcommand{\nodes}{\checkNotation{\mathcal{N}}}
\newcommand{\edges}{\checkNotation{\mathcal{E}}}
\newcommand{\greedy}[1]{\checkNotation{\overrightarrow{#1}}}
\newcommand{\kbest}[2]{\checkNotation{#2^{(#1)}}}
\newcommand{\kbestrc}[2]{\checkNotation{#2^{[#1]}}}
\renewcommand{\root}{\checkNotation{\rho}}
\newcommand{\arbs}[1]{\checkNotation{\mathcal{A}(#1)}}
\newcommand{\arbsrc}[1]{\checkNotation{\mathcal{D}(#1)}}
\newcommand{\cg}{\checkNotation{C}}
\newcommand{\cc}{\checkNotation{c}}
\newcommand{\first}[1]{\kbest{1}{#1}}
\newcommand{\second}[1]{\kbest{2}{#1}}
\newcommand{\edge}[2]{\checkNotation{(#1\,{\rightarrow}\,#2)}}
\newcommand{\edgeij}[0]{\edge{i}{j}}

\newcommand{\tree}{\checkNotation{d}}
\newcommand{\treep}{\checkNotation{\tree'}}

\newcommand{\prov}[0]{\checkNotation{\pi}}
\newcommand{\stitch}[2]{\checkNotation{#1 \looparrowright #2}}
\newcommand{\weight}[1]{\checkNotation{w\!\left(#1\right)}}
\newcommand{\edgedelete}[2]{\checkNotation{#1 - #2}}
\newcommand{\edgeinclude}[2]{\checkNotation{#1 + #2}}
\newcommand{\contract}[2]{\checkNotation{{#1}_{\!/{#2}}}}
\newcommand{\blueedges}[3]{\checkNotation{{\color{mediumblue}\textbf{b}(#1, #2, #3)}}}
\newcommand{\rededges}[3]{\checkNotation{{\color{crimson}\textbf{r}(#1, #2, #3)}}}
\newcommand{\swapcost}[2]{\checkNotation{\bar{w}_{#1}\!\left(#2\right)}}

\newcommand{\ar}{arborescence\xspace}

\newcommand{\Naively}{Na{\"i}vely\xspace}

\newcommand{\algFace}[1]{\texttt{#1}}
\newcommand{\algCall}[2]{#1\!\left( #2 \right)}

\newcommand{\opt}{\algFace{opt}}
\newcommand{\optCall}[1]{\algCall{\opt}{#1}}
\newcommand{\constrain}{\algFace{constrain}}
\newcommand{\constrainCall}[1]{\algCall{\constrain}{#1}}

\newcommand{\optnext}{\algFace{next}}
\newcommand{\optnextCall}[1]{\algCall{\optnext\!}{#1}}

\newcommand{\kbestalg}{\algFace{kbest}}
\newcommand{\kbestalgCall}[2]{\algCall{\kbestalg}{#1,#2}}

\newcommand{\kbestrcalg}{\algFace{kbest\_dep}}
\newcommand{\kbestrcalgCall}[2]{\algCall{\kbestrcalg}{#1,#2}}

\newcommand*\nodeId[1]{\tikz[baseline]{
            \node[shape=circle,draw,inner sep=1pt, text centered, text depth=0.2mm] () at (0,0.07) {\tiny $#1$};}}

\title{On Finding the $\nK$-best Non-projective Dependency Trees}

\newcommand{\ucambridge}{\emoji[twitter]{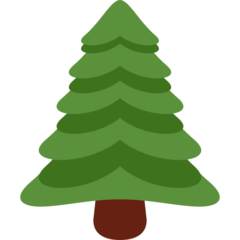}}
\newcommand{\ethz}{\emoji[twitter]{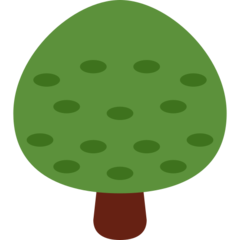}}
\newcommand{\jhu}{\emoji[twitter]{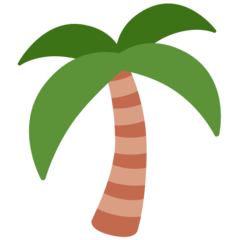}}

\author{
{Ran Zmigrod\raise1.0ex\hbox{\normalfont\ucambridge}\raise1.0ex\hbox{\normalfont}}~\;~Tim Vieira\raise1.0ex\hbox{\normalfont\jhu}~\;~Ryan Cotterell\raise1.0ex\hbox{\normalfont\ucambridge,\ethz}
\\
  \raise1.0ex\hbox{\normalfont\ucambridge}University of Cambridge~\;~\raise1.0ex\hbox{\normalfont\jhu}Johns Hopkins University~\;~\raise1.0ex\hbox{\normalfont\ethz}ETH Z\"{u}rich \\
  \texttt{rz279@cam.ac.uk}~\;~\texttt{tim.f.vieira@gmail.com} \\ \texttt{ryan.cotterell@inf.ethz.ch}
}

\date{}

\allowdisplaybreaks
\begin{document}
\maketitle
\begin{abstract}
The connection between the maximum spanning tree in a directed graph and the best dependency tree of a sentence has been exploited by the NLP community.
However, for many dependency parsing schemes, an important detail of this approach is that the spanning tree must have exactly one edge emanating from the root.
While work has been done to efficiently solve this problem for finding the one-best dependency tree, no research has attempted to extend this solution to finding the $\nK$-best dependency trees.
This is arguably a more important extension as a larger proportion of decoded trees will not be subject to the root constraint of dependency trees.
Indeed, we show that the rate of root constraint violations increases by an average of $13$ times when decoding with $\nK\!=\!50$ as opposed to $\nK\!=\!1$.
In this paper, we provide a simplification of the $\nK$-best spanning tree algorithm of \citet{CameriniFM80}.
Our simplification allows us to obtain a constant time speed-up over the original algorithm.
Furthermore,
we present a novel extension of the algorithm for decoding the $\nK$-best dependency trees of a graph which are subject to a root constraint.\footnote{\label{footnote:code}Our implementation is available at \url{https://github.com/rycolab/spanningtrees}.}
\end{abstract}

\section{Introduction}
Non-projective, graph-based dependency parsers are widespread in the NLP literature. \citep{mcdonald-etal-2005-non, dozat, stanza}.
However, despite the prevalence of $\nK$-best dependency parsing for other parsing formalisms---often in the context of re-ranking \citep{collins-koo-2005-discriminative,sangati-etal-2009-generative, zhu-etal-2015-ranking, do-rehbein-2020-neural} and other areas of NLP \citep{shen-etal-2004-discriminative, huang-chiang-2005-better, pauls-klein-2009-k, zhang-etal-2009-k}, we have only found three works that consider $\nK$-best non-projective dependency parsing \citep{hall-2007-k, hall-etal-2007-log, agic-2012-k}.
All three papers utilize the $\nK$-best spanning tree algorithm of \citet{CameriniFM80}.
Despite the general utility of $\nK$-best methods in NLP, we suspect that the relative lack of interest in $\nK$-best non-projective dependency parsing is due to the implementation complexity and nuances of \citet{CameriniFM80}'s algorithm.\footnote{In fact, an anonymous reviewer called it ``one of the most `feared' algorithms in dependency parsing.''\looseness=-1}

\begin{figure}[t]
    \centering
    \includegraphics[width=0.47\textwidth]{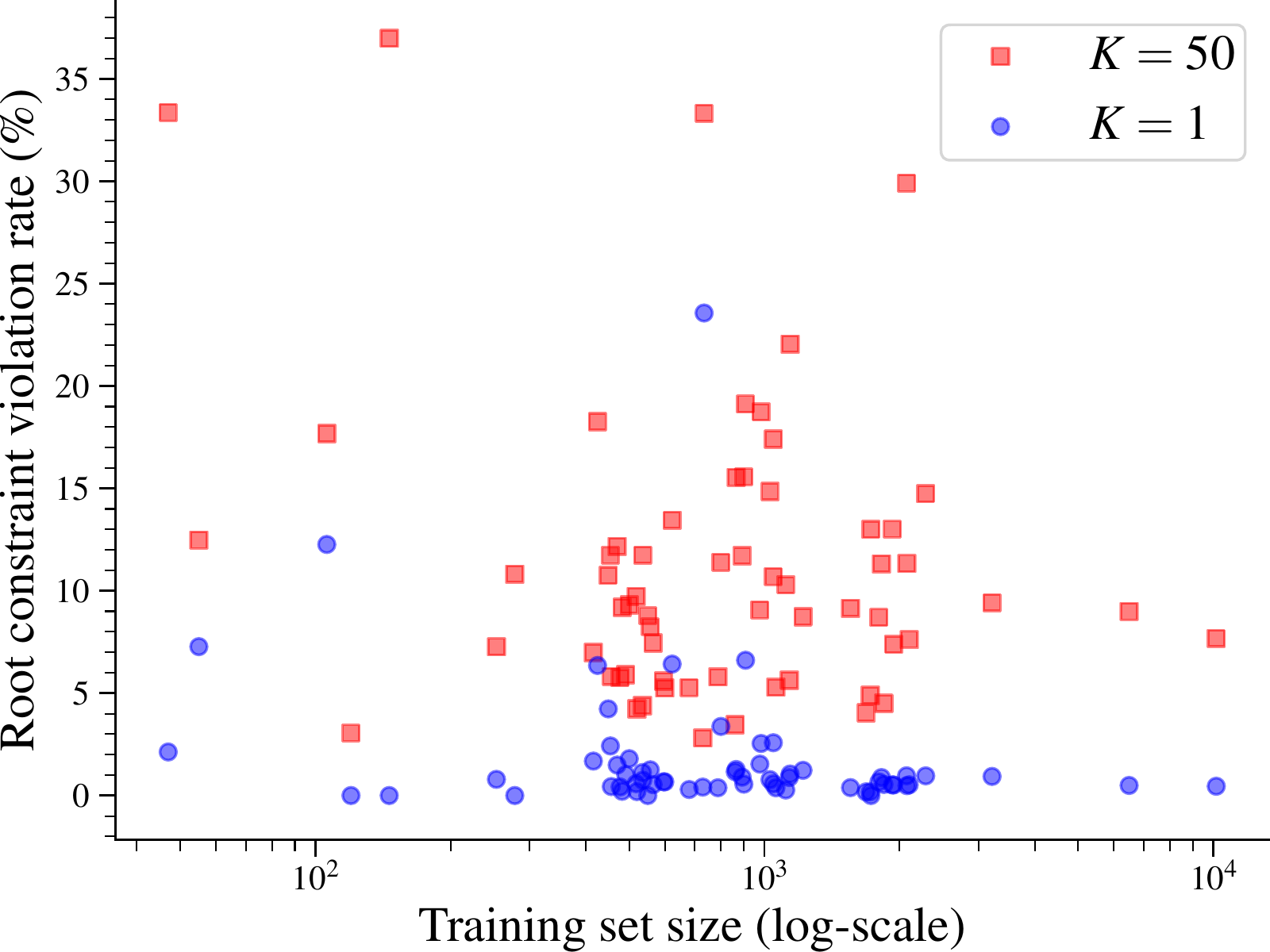}
    \caption{Violation rate of the root constraint when using regular $K$-best decoding \citep{CameriniFM80} on pre-trained models of \citet{stanza} for languages with varying training set sizes.}
    \label{fig:experiment}
\end{figure}

\newcommand{\maybetext}[1]{{\color{red}\textbf{#1}}\xspace}
We make a few changes to \citet{CameriniFM80}'s algorithm, 
which result in both a simpler algorithm and simpler proof of correctness.\footnote{While our algorithm is by no means \emph{simple},
an anonymous reviewer called it ``a big step in that direction.''}
Firstly, both algorithms follow the key property that we can find the second-best tree of a graph by removing a single edge from the graph (\cref{thm:second}); this property is used iteratively to enumerate the $\nK$-best trees in order.
Our approach to finding the second-best tree (see \cref{sec:next-best}) is faster because of it performs half as many of the expensive cycle-contraction operations (see \cref{sec:best}). Overall, this change is responsible for our $1.39$x speed-up (see \cref{sec:kbest}).
Secondly, their proof of correctness is based on reasoning about a complicated ordering on the edges in the $\nK\th$ tree \citep[Section 4]{CameriniFM80};
our proof side-steps the complicated ordering by directly reasoning over the ancestry relations of the $\nK\th$ tree.
Consequently, our proofs of correctness are considerably simpler and shorter.
Throughout the paper, we provide the statements of all lemmas and theorems in the main text, but defer all proofs to the appendix.

In addition to simplifying \citet{CameriniFM80}'s algorithm,
we offer a novel extension.
For many dependency parsing schemes such as the Universal Dependency (UD) scheme \citep{ud}, there is a restriction on dependency trees to only have one edge emanate from the root.\footnote{There are certain exceptions to this such as the Prague Treebank \citep{prague_dep}.}
Finding the maximally weighted spanning tree that obeys this constraint was considered by \citet{GabowT84} who extended the $\bigo{\nN^2}$ maximum spanning tree algorithm of \citet{Tarjan77, camerini1979note}.
However, no algorithm exists for $\nK$-best decoding of dependency trees subject to a root constraint.
As such, we provide the first $\nK$-best algorithm that returns dependency trees that obey the root constraint.

To motivate the practical necessity of our extension, consider \cref{fig:experiment}.
\cref{fig:experiment} shows the percentage of trees that violate the root constraint when doing one-best and $50$-best decoding for 63 languages from the UD treebank \citep{ud} using the pre-trained model of \citet{stanza}.\footnote{\citet{zmigrod-etal-2020-please} conduct a similar experiment for only the one-best tree.}$^,$\footnote{We note that \citet{stanza} do apply the root constraint for one-best decoding, albeit with a sub-optimal algorithm.}
We find that decoding without the root constraint has a much more extreme effect when decoding the $50$-best than the one-best.
Specifically, we observe that on average, the number of violations of the root constraint increased by $13$ times, with the worst increase being 44 times.
The results thus suggest that finding $\nK$-best trees that obey the root constraint from a non-projective dependency parser requires a specialist algorithm.
We provide a more detailed results table in \cref{app:table}, including root constraint violation rates for $\nK\!=\!5$, $\nK\!=\!10$, and $\nK\!=\!20$.
Furthermore, we note that the $\nK$-best algorithm may also be used for marginalization of latent variables \citep{correiaNAM20} and for constructing parsers with global scoring functions \citep{leeLZ16}.
\begin{toappendix}
\subsection*{Results Table for \cref{fig:experiment}}\label{app:table}
\begin{center}
    \resizebox{!}{0.47\textheight}{
    \begin{tabular}{lccccccc}
          \multirow{2}{*}{\bf Language} & \multirow{2}{*}{\bf $\abs{\text{Train}}$} & \multirow{2}{*}{\bf $\abs{\text{Test}}$} & \multicolumn{5}{c}{\bf Root Constraint Violation Rate ($\%$)} \\
          & &  & $\nK=1$ & $\nK=5$  & $\nK=10$ & $\nK=20$ & $\nK=50$ \\
          \midrule
Czech & 68495 & 10148 & 0.45 & 5.07 & 6.18 & 6.76 & 7.67 \\
Russian & 48814 & 6491 & 0.49 & 5.07 & 6.58 & 7.66 & 8.99 \\
Estonian & 24633 & 3214 & 0.93 & 5.59 & 7.02 & 8.24 & 9.42 \\
Korean & 23010 & 2287 & 0.96 & 6.68 & 9.51 & 11.91 & 14.74 \\
Latin & 16809 & 2101 & 0.52 & 5.17 & 5.57 & 6.25 & 7.62 \\
Norwegian & 15696 & 1939 & 0.52 & 4.26 & 5.20 & 6.22 & 7.38 \\
Ancient Greek & 15014 & 1047 & 0.57 & 4.74 & 7.00 & 8.38 & 10.69 \\
French & 14450 & 416 & 1.68 & 3.85 & 4.95 & 5.81 & 6.98 \\
Spanish & 14305 & 1721 & 0.17 & 2.25 & 3.25 & 3.96 & 4.89 \\
Old French & 13909 & 1927 & 0.52 & 6.81 & 9.41 & 11.38 & 13.01 \\
German & 13814 & 977 & 1.54 & 5.12 & 6.37 & 7.63 & 9.06 \\
Polish & 13774 & 1727 & 0.00 & 4.76 & 7.86 & 10.11 & 13.00 \\
Hindi & 13304 & 1684 & 0.18 & 1.34 & 2.19 & 2.98 & 4.04 \\
Catalan & 13123 & 1846 & 0.54 & 2.32 & 2.97 & 3.68 & 4.51 \\
Italian & 13121 & 482 & 0.21 & 4.02 & 5.66 & 7.25 & 9.19 \\
English & 12543 & 2077 & 0.48 & 9.12 & 10.73 & 11.12 & 11.34 \\
Dutch & 12264 & 596 & 0.67 & 3.39 & 4.18 & 4.82 & 5.59 \\
Finnish & 12217 & 1555 & 0.39 & 4.72 & 6.12 & 7.39 & 9.15 \\
Classical Chinese & 11004 & 2073 & 0.96 & 22.52 & 25.95 & 28.09 & 29.91 \\
Latvian & 10156 & 1823 & 0.88 & 7.05 & 8.77 & 9.95 & 11.31 \\
Bulgarian & 8907 & 1116 & 0.27 & 4.66 & 6.73 & 8.16 & 10.29 \\
Slovak & 8483 & 1061 & 0.38 & 4.81 & 5.34 & 5.29 & 5.29 \\
Portuguese & 8328 & 477 & 0.42 & 3.31 & 4.15 & 4.76 & 5.75 \\
Romanian & 8043 & 729 & 0.41 & 1.26 & 1.66 & 2.16 & 2.81 \\
Japanese & 7125 & 550 & 0.00 & 5.13 & 6.24 & 7.20 & 8.79 \\
Croatian & 6914 & 1136 & 0.88 & 2.90 & 3.71 & 4.44 & 5.62 \\
Slovenian & 6478 & 788 & 0.38 & 2.66 & 3.53 & 4.59 & 5.79 \\
Arabic & 6075 & 680 & 0.29 & 3.79 & 4.15 & 4.72 & 5.27 \\
Ukrainian & 5496 & 892 & 0.90 & 7.49 & 9.15 & 10.13 & 11.72 \\
Basque & 5396 & 1799 & 0.67 & 3.64 & 5.06 & 6.67 & 8.71 \\
Hebrew & 5241 & 491 & 1.02 & 2.81 & 4.01 & 5.04 & 5.90 \\
Persian & 4798 & 600 & 0.67 & 2.43 & 3.47 & 4.28 & 5.25 \\
Indonesian & 4477 & 557 & 1.26 & 4.06 & 5.48 & 6.65 & 8.25 \\
Danish & 4383 & 565 & 0.53 & 4.35 & 5.59 & 6.35 & 7.45 \\
Swedish & 4303 & 1219 & 1.23 & 4.63 & 6.08 & 7.09 & 8.73 \\
Old Church Slavonic & 4124 & 1141 & 1.05 & 14.32 & 17.64 & 19.88 & 22.05 \\
Urdu & 4043 & 535 & 1.12 & 2.47 & 3.08 & 3.60 & 4.39 \\
Chinese & 3997 & 500 & 1.80 & 4.80 & 5.90 & 7.68 & 9.31 \\
Turkish & 3664 & 983 & 2.54 & 12.47 & 15.53 & 17.09 & 18.73 \\
Gothic & 3387 & 1029 & 0.78 & 8.65 & 11.18 & 13.10 & 14.86 \\
Serbian & 3328 & 520 & 0.19 & 2.04 & 2.60 & 3.16 & 4.23 \\
Galician & 2272 & 861 & 1.16 & 2.07 & 2.36 & 2.88 & 3.46 \\
North Sami & 2257 & 865 & 1.27 & 7.49 & 10.15 & 12.43 & 15.54 \\
Armenian & 1975 & 278 & 0.00 & 7.34 & 8.42 & 9.64 & 10.81 \\
Greek & 1662 & 456 & 0.44 & 3.20 & 4.19 & 4.80 & 5.82 \\
Uyghur & 1656 & 900 & 0.56 & 7.18 & 9.64 & 12.24 & 15.57 \\
Vietnamese & 1400 & 800 & 3.38 & 6.78 & 8.25 & 9.56 & 11.39 \\
Afrikaans & 1315 & 425 & 6.35 & 13.65 & 14.73 & 16.12 & 18.26 \\
Wolof & 1188 & 470 & 1.49 & 6.89 & 8.32 & 9.91 & 12.17 \\
Maltese & 1123 & 518 & 0.58 & 5.17 & 6.70 & 8.12 & 9.73 \\
Telugu & 1051 & 146 & 0.00 & 27.81 & 32.81 & 36.16 & 36.99 \\
Scottish Gaelic & 1015 & 536 & 0.75 & 7.16 & 8.97 & 10.20 & 11.75 \\
Hungarian & 910 & 449 & 4.23 & 7.44 & 8.66 & 9.82 & 10.75 \\
Irish & 858 & 454 & 2.42 & 7.14 & 8.68 & 10.23 & 11.73 \\
Tamil & 400 & 120 & 0.00 & 1.17 & 1.50 & 1.83 & 3.05 \\
Marathi & 373 & 47 & 2.13 & 20.85 & 21.70 & 27.34 & 33.36 \\
Belarusian & 319 & 253 & 0.79 & 5.61 & 9.05 & 8.99 & 7.27 \\
Lithuanian & 153 & 55 & 7.27 & 9.82 & 10.36 & 10.82 & 12.47 \\
Kazakh & 31 & 1047 & 2.58 & 7.97 & 10.68 & 13.45 & 17.41 \\
Upper Sorbian & 23 & 623 & 6.42 & 9.34 & 10.72 & 11.78 & 13.45 \\
Kurmanji & 20 & 734 & 23.57 & 27.06 & 29.22 & 30.87 & 33.33 \\
Buryat & 19 & 908 & 6.61 & 10.37 & 13.00 & 15.48 & 19.13 \\
Livvi & 19 & 106 & 12.26 & 14.15 & 15.00 & 15.99 & 17.68 \\
\end{tabular}
}
\end{center}

\end{toappendix}

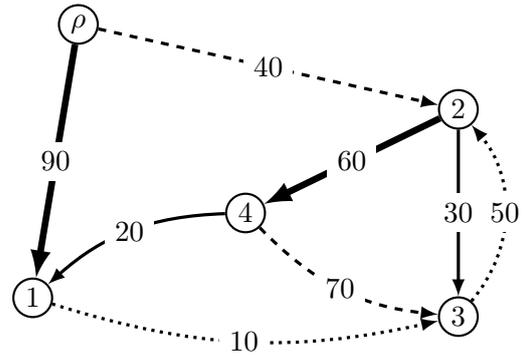
\begin{figure}[t]
    \centering
    \begin{tikzpicture}
\begin{scope}[every node/.style={circle,thick,draw, inner sep=2pt}]
    \node (2) at (0.8, 1.374) {$2$};
    \node (3) at (0.8, -1.374) {$3$};
    \node (4) at (-2, 0) {$4$};
    \node (1) at (-4.8, -1.125) {$1$};
    \node (r) at (-4.2, 2.5) {$\root$};
\end{scope}

\begin{scope}[>=latex,
              every node/.style={fill=white},
              every edge/.style={draw, very thick}]
    \path [->] (r) edge[line width=2.5pt] node {$90$} (1);
    \path [->] (r) edge[dashed] node {$40$} (2);
    \path [->] (1) edge[bend right=15, dotted,] node {$10$} (3);
    \path [->] (2) edge[line width=2.5pt] node {$60$} (4);
    \path [->] (2) edge[] node {$30$} (3);
    \path [->] (3) edge[bend right=40, dotted] node {$50$} (2);
    \path [->] (4) edge[bend right=20, dashed] node {$70$} (3);
    \path [->] (4) edge[bend right=20] node {$20$} (1);
\end{scope}

\end{tikzpicture}
    \caption{Example graph $\graph$ (taken from \citet{zmigrod-etal-2020-please}). Edges that are part of both the best tree $\first{\graph}$ and the best dependency tree $\kbestrc{1}{\graph}$ are marked as thick solid edges. Edges only in $\first{\graph}$ are dashed and edges only in $\kbestrc{1}{\graph}$ are dotted.}
    \label{fig:example}
\end{figure}

\section{Finding the Best Tree}\label{sec:best}
We consider the study of \defn{rooted directed weighted graphs}, which we will abbreviate to simply \defn{graphs}.\footnote{As we use the algorithm in \citet{zmigrod-etal-2020-please} as our base algorithm, we borrow their notation wherever convenient.}
A graph is given by $\graph=(\root,\nodes, \edges)$ where $\nodes$ is a set of $\nN+1$ nodes with a designated root node $\root\in \nodes$ and $\edges$ is a set of directed weighted edges. Each edge $e=\edgeij\in \edges$ has a weight $\weight{e}\in\mathbb{R}_{+}$.
We assume that self-loops are not allowed in the graph (i.e., $\edge{i}{i}\not\in \edges$).
Additionally, we assume our graph is not a multi-graph, therefore, there can exist at most one edge from node $i$ to node $j$.\footnote{We make this assumption for simplicity, the algorithms presented here will also work with multi-graphs. This might be desirable for decoding labeled dependency trees. However, we note that in most graph-based parsers such as \citet{stanza} and \citet{ma}, dependency labels are extracted after the unlabeled tree has been decoded.}
When it is clear from context, we abuse notation and use $j\in\graph$ and $e\in\graph$ for $j\in \nodes$ and $e\in \edges$ respectively.
When discussing runtimes, we will assume a fully connected graph ($\abs{\edges}=\nN^2$).\footnote{We make this assumption as in the context of dependency parsing, we generate scores for each possible edge. Furthermore, \cite{Tarjan77} prove that the runtime of finding the best tree for dense graphs is $\bigo{\nN^2}$. This is $\bigo{\abs{\edges}\log \nN}$ in the non-dense case.}
An \defn{\ar} (henceforth called a \defn{tree}) of $\graph$ is a subgraph $\tree=(\root,\nodes,\edges')$ such that $\edges'\subseteq \edges$ and the following is true:
\begin{enumerate}[leftmargin=2.5em, topsep=8pt, itemsep=1pt, parsep=5pt]
\item For all $j\in \nodes \setminus \{\root\}$, $\abs{\{\edge{\_}{j}\in \edges'\}}=1$.
\item $\tree$ does not contain any cycles.
\end{enumerate}
Other definitions of trees can also include that there is at least one edge emanating from the root.
However, this condition is immediately satisfied by the above two conditions.
A \defn{dependency tree} $\tree=(\root,\nodes,\edges')$ is a tree with the extra constraint
\begin{enumerate}[leftmargin=2.5em, topsep=8pt, itemsep=1pt, parsep=5pt]
\item[3.] $\abs{\{ \edge{\root}{\_}\in \nodes' \}}= 1$
\end{enumerate}
The set of all trees and dependency trees in a graph are given by $\arbs{\graph}$ and $\arbsrc{\graph}$ respectively.
The weight of a tree is given by the sum of its edge weights\footnote{For inference, the weight of a trees often decomposes multiplicatively rather than additively over the edges. One can take the exponent (or logarithm) of the original edge weights to make the weights distribute additively (or multiplicative).}
\begin{equation}
    \weight{\tree} = \sum_{e\in\tree} \weight{e}
\end{equation}

This paper concerns finding the $\nK$ highest-weighted (henceforce called \defn{$\boldsymbol{\nK}$-best}) tree or dependency tree, these are denoted by $\kbest{K}{\graph}$ and $\kbestrc{K}{\graph}$ respectively.
\citet{Tarjan77, camerini1979note} provided the details for an $\bigo{\nN^2}$ algorithm for decoding the one-best tree.
This algorithm was extended by \citet{GabowT84} to find the best dependency tree in $\bigo{\nN^2}$ time.
We borrow the algorithm (and notation) of \citet{zmigrod-etal-2020-please}, who provide an exposition and proofs of these algorithms in the context of non-projective dependency parsing.
The pseudocode for finding $\first{\graph}$ and $\kbestrc{1}{\graph}$ is given in \cref{alg:ma}.
We briefly describe the key components of the algorithm.\footnote{For a more complete and detailed description as well as a proof of correctness, please refer to the original manuscripts.}

The \defn{greedy graph} of $\graph$ is denoted by $\greedy{\graph}=(\root,\nodes,\edges')$ where $\edges'$ contains the highest weighted incoming edge to each non-root node.
Therefore, if $\greedy{\graph}$ has no cycles, then $\greedy{\graph}=\first{\graph}$.
A cycle $\cg$ in $\greedy{\graph}$ is called a \defn{critical cycle}.
If we encounter a critical cycle in the algorithm, we contract the graph by the critical cycle.
A graph \defn{contraction}, $\contract{\graph}{\cg}$, by a cycle $\cg$ replaces the nodes in $\cg$ by a mega-node $\cc$ such that the nodes of $\contract{\graph}{\cg}$ are $\nodes\setminus \cg \cup \{\cc\}$.
Furthermore, for each edge $e=\edgeij\in\graph$:
\begin{enumerate}[leftmargin=2.5em, topsep=8pt, itemsep=1pt, parsep=5pt]
\item If $i\not\in\cg$ and $j\in\cg$, then $e'=\edge{i}{c}\in\contract{\graph}{\cg}$ such that $\weight{e'}=\weight{e} + \weight{\greedy{\cg_j}}$ where $\cg_j$ is the subgraph of $\cg$ rooted at $j$.
\item If $i\in\cg$ and $j\not\in\cg$, then $e'=\edge{c}{j}\in\contract{\graph}{\cg}$ such that $\weight{e'}=\weight{e}$.
\item If $i\not\in\cg$ and $j\not\in\cg$, then $e\in\contract{\graph}{\cg}$.
\item If $i\in\cg$ and $j\in\cg$, then there is no edge related to $\edgeij$ in $\contract{\graph}{\cg}$.
\end{enumerate}
There also exists a bookkeeping function $\prov$ such that for all $e'\in\contract{\graph}{\cg}$, $\prov(e')\in\graph$.
This bookkeeping function returns the edge in the original graph that led to the creation of the edge in the contracted graph using one of the constructions above.

\begin{figure}[t]
\input{figures/mst_alg}
\caption{Algorithms for finding $\first{\graph}$ and $\kbestrc{1}{\graph}$. These are from \citet{zmigrod-etal-2020-please}.}
\label{alg:ma}
\end{figure}

Finding $\first{\graph}$ is then the task of finding a contracted graph $\graph'$ such that $\greedy{\graph'}=\first{\graph'}$. Once this is done, we can stitch back the cycles we contracted.
If $\graph'=\contract{\graph}{\cg}$, for any $\tree\in\arbs{\contract{\graph}{\cg}}$, $\stitch{\tree}{\cg}\in\arbs{\graph}$ is the tree made with edges $\prov(\tree)$ ($\prov$ applied to each edge $\tree$) and $\greedy{\cg_j}$ where $\cg_j$ is the subgraph of the nodes in $\cg$ rooted at node $j$ and $\prov(e)=\edge{i}{j}$ for $e=\edge{i}{\cc}\in\tree$.
The contraction weighting scheme means that $\weight{\tree}=\weight{\stitch{\tree}{\cg}}$ \citep{georgiadis}.
Therefore, $\first{\graph}=\first{(\stitch{\first{\graph'}}{\cg})}$.

The strategy for finding $\kbestrc{1}{\graph}$ is to find the contracted graph for $\first{\graph}$ and attempt to remove edges emanating from the root. This was first proposed by \citet{GabowT84}. When we consider removing an edge emanating from the root, we are doing this in a possibly contracted graph, and so an edge $\edge{\root}{j}$ may exist multiple times in the graph. We denote $\medgedelete{\graph}{e}$ to be the graph $\graph$ with all edges with the same end-points as $e$ removed.
\cref{fig:example} gives an example of a graph $\graph$, its best tree $\first{\graph}$, and its best dependency tree $\kbestrc{1}{\graph}$.

\begin{figure*}[t!]
    \centering
    \tikzset{cross/.style={cross out, draw=black, minimum size=2*(#1-\pgflinewidth), inner sep=0pt, outer sep=0pt},
cross/.default={2pt}}
\begin{subfigure}[b]{.3\linewidth}
\begin{tikzpicture}
\begin{scope}[every node/.style={circle,thick,draw, inner sep=2pt}]
    \node (2) at (3.36, 0.916) {$2$};
    \node (3) at (3.36, -0.916) {$3$};
    \node (4) at (1.68, 0) {$4$};
    \node (1) at (0, -0.75) {$1$};
    \node (r) at (0.36, 1.6) {$\root$};
\end{scope}
\begin{scope}[>=latex,
              every node/.style={fill=white},
              every edge/.style={draw, very thick}]
    \path [->] (r) edge[] (1);
    \path [->] (r) edge[] (2);
    \path [->] (r) edge[dashed, bend right=45] (3);
    \path [->] (r) edge[dashed, ] (4); 
    
    \path [->] (1) edge[dashed, bend right=30] (2);
    \path [->] (1) edge[dashed, bend right=15] (3);
    \path [->] (1) edge[dashed] (4); 
    
    \path [->] (2) edge[dashed, bend right=30] (1);
    \path [->] (2) edge[dashed] (3);
    \path [->] (2) edge[] (4); 
    
    \path [->] (3) edge[dashed, bend left=30] (1);
    \path [->] (3) edge[dashed, bend right=40] (2);
    \path [->] (3) edge[dashed, bend right=20] (4); 
    
    \path [->] (4) edge[dashed, bend right=20] (1);
    \path [->] (4) edge[dashed, bend right=20] (2);
    \path [->] (4) edge[bend right=20] (3); 
\end{scope}
\begin{scope}[every node/.style={ellipse, fill=white, line width=0pt, fill opacity=0}]
\node[minimum height=3.3cm, minimum width=2.1cm] (blue) at (0.25, 0.4) {};
\node[minimum height=1.3cm, minimum width=2.1cm] (red) at (2.8, -0.65) {};
\end{scope}
\node (blueLab) at (1.7, 1.9) {\color{mediumblue} \small $\hphantom{\blueedges{\graph}{e}{\first{\graph}}}$};
\node (redLab) at (1.55, -1.4) {\color{crimson} \small $\hphantom{\rededges{\graph}{e}{\first{\graph}}}$};
\end{tikzpicture}
\caption{}
\label{subfig:a}
\end{subfigure}
\begin{subfigure}[b]{.3\linewidth}
\begin{tikzpicture}
\begin{scope}[every node/.style={circle,thick,draw, inner sep=2pt}]
    \node (2) at (3.36, 0.916) {$2$};
    \node (3) at (3.36, -0.916) {$3$};
    \node (4) at (1.68, 0) {$4$};
    \node (1) at (0, -0.75) {$1$};
    \node (r) at (0.36, 1.6) {$\root$};
\end{scope}
\begin{scope}[>=latex,
              every node/.style={fill=white},
              every edge/.style={draw, very thick}]
    \path [->] (r) edge[] (1);
    \path [->] (r) edge[] (2);
    \path [->] (r) edge[dashed] (4); 
    
    \path [->] (1) edge[dashed] (4); 
    
    \path [->] (2) edge[] node[inner sep=1pt] {$e$} (4); 
    
    \path [->] (3) edge[dashed, bend right=20] (4); 
    
    \path [->] (4) edge[bend right=20] (3); 
\end{scope}
\begin{scope}[every node/.style={ellipse, fill, line width=0pt, fill opacity=0.15}]
\node[minimum height=3.3cm, minimum width=2.1cm, fill=blue] (blue) at (0.25, 0.4) {};
\node[minimum height=1.3cm, minimum width=2.1cm, fill=red] (red) at (2.8, -0.65) {};
\end{scope}
\node (blueLab) at (1.7, 1.9) {\color{mediumblue} \small $\blueedges{\graph}{e}{\first{\graph}}$};
\node (redLab) at (1.55, -1.4) {\color{crimson} \small $\rededges{\graph}{e}{\first{\graph}}$};
\end{tikzpicture}
\caption{}
\label{subfig:b}
\end{subfigure}
\begin{subfigure}[b]{.3\linewidth}
\begin{tikzpicture}
\begin{scope}[every node/.style={circle,thick,draw, inner sep=2pt}]
    \node (2) at (3.36, 0.916) {$2$};
    \node (3) at (3.36, -0.916) {$3$};
    \node (4) at (1.68, 0) {$4$};
    \node (1) at (0, -0.75) {$1$};
    \node (r) at (0.36, 1.6) {$\root$};
\end{scope}
\begin{scope}[>=latex,
              every node/.style={fill=white},
              every edge/.style={draw, very thick}]
    \path [->] (r) edge[] (1);
    \path [->] (r) edge[] (2);
    \path [->] (r) edge[dashed] (4); 
    
    \path [->] (1) edge[] node[inner sep=1pt] {$e'$} (4); 

    \path [->] (2) edge[\Cdead] node[cross, minimum size=10pt, line width=5pt] {} (4); 

    \path [->] (3) edge[dashed, bend right=20] (4); 

    \path [->] (4) edge[bend right=20] (3); 
\end{scope}
\begin{scope}[every node/.style={ellipse, fill, line width=0pt, fill opacity=0.15}]
\node[minimum height=3.3cm, minimum width=2.1cm, fill=blue] (blue) at (0.25, 0.4) {};
\node[minimum height=1.3cm, minimum width=2.1cm, fill=red] (red) at (2.8, -0.65) {};
\end{scope}
\node (blueLab) at (1.7, 1.9) {\color{mediumblue} \small $\blueedges{\graph}{e}{\first{\graph}}$};
\node (redLab) at (1.55, -1.4) {\color{crimson} \small $\rededges{\graph}{e}{\first{\graph}}$};
\end{tikzpicture}
\caption{}
\label{subfig:c}
\end{subfigure}

\begin{subfigure}[b]{.3\linewidth}
\begin{tikzpicture}
\begin{scope}[every node/.style={circle,thick,draw, inner sep=2pt}]
    \node (2) at (3.36, 0.916) {$2$};
    \node (3) at (3.36, -0.916) {$3$};
    \node (4) at (1.68, 0) {$4$};
    \node (1) at (0, -0.75) {$1$};
    \node (r) at (0.36, 1.6) {$\root$};
\end{scope}
\begin{scope}[>=latex,
              every node/.style={fill=white},
              every edge/.style={draw, very thick}]
    \path [->] (r) edge[] (1);
    \path [->] (r) edge[] (2);
    \path [->] (r) edge[dashed] (4); 

    \path [->] (1) edge[dashed] (4); 

    \path [->] (2) edge[\Cdead] node[cross, minimum size=10pt, line width=5pt] {} (4); 

    \path [->] (3) edge[bend right=20] node[inner sep=0pt] {$e''$} (4); 

    \path [->] (4) edge[bend right=20] node[inner sep=1pt] {$f$} (3); 
\end{scope}
\begin{scope}[every node/.style={ellipse, fill, line width=0pt, fill opacity=0.15}]
\node[minimum height=3.3cm, minimum width=2.1cm, fill=blue] (blue) at (0.25, 0.4) {};
\node[minimum height=1.3cm, minimum width=2.1cm, fill=red] (red) at (2.8, -0.65) {};
\end{scope}
\node (blueLab) at (1.7, 1.9) {\color{mediumblue} \small $\blueedges{\graph}{e}{\first{\graph}}$};
\node (redLab) at (1.55, -1.4) {\color{crimson} \small $\rededges{\graph}{e}{\first{\graph}}$};
\end{tikzpicture}
\caption{}
\label{subfig:d}
\end{subfigure}
\begin{subfigure}[b]{.3\linewidth}
\begin{tikzpicture}
\begin{scope}[every node/.style={circle,thick,draw, inner sep=2pt}]
    \node (2) at (3.36, 0.916) {$2$};
    \node (3) at (3.36, -0.916) {$3$};
    \node (4) at (1.68, 0) {$4$};
    \node (1) at (0, -0.75) {$1$};
    \node (r) at (0.36, 1.6) {$\root$};
\end{scope}
\begin{scope}[>=latex,
              every node/.style={fill=white},
              every edge/.style={draw, very thick}]
    \path [->] (r) edge[] (1);
    \path [->] (r) edge[] (2);
    \path [->] (r) edge[dashed] (4); 

    \path [->] (1) edge[dashed] (4); 

    \path [->] (2) edge[] node[inner sep=1pt] {$f'$} (3);
    \path [->] (2) edge[\Cdead] node[cross, minimum size=10pt, line width=5pt] {} (4); 

    \path [->] (3) edge[bend right=20] node[inner sep=0pt] {$e''$} (4); 

    \path [->] (4) edge[dashed, bend right=20] node[inner sep=1pt] {$f$} (3); 
\end{scope}
\begin{scope}[every node/.style={ellipse, fill, line width=0pt, fill opacity=0.15}]
\node[minimum height=3.3cm, minimum width=2.1cm, fill=blue] (blue) at (0.25, 0.4) {};
\node[minimum height=1.3cm, minimum width=2.1cm, fill=red] (red) at (2.8, -0.65) {};
\end{scope}
\node (blueLab) at (1.7, 1.9) {\color{mediumblue} \small $\blueedges{\graph}{e}{\first{\graph}}$};
\node (redLab) at (1.55, -1.4) {\color{crimson} \small $\rededges{\graph}{e}{\first{\graph}}$};
\end{tikzpicture}
\caption{}
\label{subfig:e}
\end{subfigure}
\begin{subfigure}[b]{.3\linewidth}
\begin{tikzpicture}
\begin{scope}[every node/.style={circle,thick,draw, inner sep=2pt}]
    \node (2) at (3.36, 0.916) {$2$};
    \node (3) at (3.36, -0.916) {$3$};
    \node (4) at (1.68, 0) {$4$};
    \node (1) at (0, -0.75) {$1$};
    \node (r) at (0.36, 1.6) {$\root$};
\end{scope}
\begin{scope}[>=latex,
              every node/.style={fill=white},
              every edge/.style={draw, very thick}]
    \path [->] (r) edge[] (1);
    \path [->] (r) edge[] (2);

    \path [->] (2) edge[] node[inner sep=1pt] {$f'$} (3);
    \path [->] (2) edge[] node[inner sep=1pt] {$e$} (4); 

    \path [->] (3) edge[dashed, bend right=20] node[inner sep=0pt] {$e''$} (4); 

    \path [->] (4) edge[dashed, bend right=20] node[inner sep=1pt] {$f$} (3); 
\end{scope}
\begin{scope}[every node/.style={ellipse, fill=white, line width=0pt, fill opacity=0}]
\node[minimum height=3.3cm, minimum width=2.1cm] (blue) at (0.25, 0.4) {};
\node[minimum height=1.3cm, minimum width=2.1cm] (red) at (2.8, -0.65) {};
\end{scope}
\node (blueLab) at (1.7, 1.9) {\color{mediumblue} \small $\hphantom{\blueedges{\graph}{e}{\first{\graph}}}$};
\node (redLab) at (1.55, -1.4) {\color{crimson} \small $\hphantom{\rededges{\graph}{e}{\first{\graph}}}$};
\end{tikzpicture}
\caption{}
\label{subfig:f}
\end{subfigure}
    \caption{
Worked example of \cref{lemma:one-diff}.
Consider a fully connected graph, $\graph$, of the example given in \cref{fig:example} as given in (a).
Suppose that the solid edges in (a) represent $\greedy{\graph}$.
Therefore, $\first{\graph}=\greedy{\graph}$.
Next, suppose that we know that $e=\edge{2}{4}\in\first{\graph}$ is not in $\second{\graph}$.
Then one of the dashed edges in (b) must be in $\second{\graph}$ as $\nodeId{4}$ must have an incoming edge.
The edges emanating from $\nodeId{\root}$ and $\nodeId{1}$ make up the set of {\color{mediumblue}blue} edges, $\blueedges{\graph}{e}{\first{\graph}}$ while the edge emanating from $\nodeId{3}$ makes the set of {\color{crimson}red} edges, $\rededges{\graph}{e}{\first{\graph}}$.
If $e'\in\blueedges{\graph}{e}{\first{\graph}}$ is in $\second{\graph}$ as in (c), then the solid lines in (c) make a tree and $\second{\graph}$ differs from $\first{\graph}$ by exactly one {\color{mediumblue}blue} edge of $e$.
Otherwise, we know that $e''\in\rededges{\graph}{e}{\first{\graph}}$ is in $\second{\graph}$ as in (d).
However, the solid edges in (d) contain a cycle between $\nodeId{3}$ and $\nodeId{4}$ with edges $e''$ and $f$.
We could break the cycle at $\nodeId{3}$ and include edge $f'$ in our tree as in (e).
However, while the solid edges in (e) make a valid tree, as $\weight{e}>\weight{e''}$ and $\weight{f}>\weight{f'}$, the tree given by the solid lines of (f) will have a higher weight.
This would mean that $e\in\second{\graph}$ which leads to a contradiction.
Therefore, we must break the cycle at $\nodeId{4}$, which leads us to a tree as in (c).
Consequently, $\second{\graph}$ will differ from $\first{\graph}$ by exactly one {\color{mediumblue}blue} edge of $e$.\looseness=-1}\label{fig:blue}
\end{figure*}

The runtime complexity of finding $\first{\graph}$ or $\kbestrc{1}{\graph}$ is $\bigo{\nN^2}$ for dense graphs by using efficient priority queues and sorting algorithms \citep{Tarjan77, GabowT84}.
We assume this runtime for the remainder of the paper.

\section{Finding the Second Best Tree}\label{sec:next-best}
In the following two sections, we provide a simplified reformulation of \citet{CameriniFM80} to find the $\nK$-best trees.
The simplifications additionally provide a constant time speed-up over \citet{CameriniFM80}'s algorithm.
We discuss the differences throughout our exposition.

The underlying concept behind finding the $\nK$-best tree, is that $\kbest{K}{\graph}$ is the second best tree $\second{\graph'}$ of some subgraph $\graph'\subseteq\graph$.
In order to explore the space of subgraphs, we introduce the concept of edge inclusion and exclusion graphs.

\begin{definition}[Edge inclusion and exclusion]
For any graph $\graph$ and edge $e\in\graph$, the \defn{edge-inclusion graph} $\edgeinclude{\graph}{e}\subset\graph$ is the graph such that for any $\tree\in\arbs{\edgeinclude{\graph}{e}}$, $e\in\tree$.
Similarly, the \defn{edge-exclusion graph} $\edgedelete{\graph}{e}\subset\graph$ is the graph such that for any $\tree\in\arbs{\edgedelete{\graph}{e}}$, $e\not\in\tree$.
\end{definition}
When we discuss finding the $\nK$-best dependency trees in \cref{sec:kbest-rc}, we implicitly change the above definition to use $\arbsrc{\edgeinclude{\graph}{e}}$ and $\arbsrc{\edgedelete{\graph}{e}}$ instead of $\arbs{\edgeinclude{\graph}{e}}$ and $\arbs{\edgedelete{\graph}{e}}$ respectively.

In this section, we will specifically focus on finding $\second{\graph}$, we extend this to finding the $\kbest{k}{\graph}$ in \cref{sec:kbest}.
Finding $\second{\graph}$ relies on the following fundamental theorem.

\begin{restatable*}[]{thm}{thmSecond}
\label{thm:second}%
For any graph $\graph$ and $e\in\first{\graph}$
\begin{equation}
    \second{\graph}=\first{(\edgedelete{\graph}{e})}
\end{equation}
where
\begin{equation}\label{eq:del-e}
    e=\argmax_{e'\in\first{\graph}}\weight{\first{(\edgedelete{\graph}{e'})}}
\end{equation}
\end{restatable*}
\begin{toappendix}
\thmSecond
\begin{proof}
There must be at least one edge $e\in\first{\graph}$ such that $e\not\in\second{\graph}$.
Therefore, there exists an $e\in\first{\graph}$ such that $\second{\graph}=\first{(\edgedelete{\graph}{e})}$.
Now suppose by way of contradiction that $e$ is not as given in \cref{eq:del-e}.
If we choose an $e'$ that satisfies \cref{eq:del-e}, then by definition $\weight{\first{(\edgedelete{\graph}{e'})}} > \weight{\first{(\edgedelete{\graph}{e})}}$.
As $\first{(\edgedelete{\graph}{e'})}\neq\first{\graph}$, we arrive at a contradiction.
\end{proof}
\end{toappendix}

\begin{figure}[t!]
\input{figures/next_alg}
\caption{Algorithm for finding $\first{\graph}$, the best edge $e$ to delete to find $\second{\graph}$, and $\weight{\second{\graph}}$.}
\label{alg:next-best}
\end{figure}

\cref{thm:second} states that we can find $\second{\graph}$ by identifying an edge $e\in\first{\graph}$ such that $\second{\graph}=\first{(\edgedelete{\graph}{e})}$.
We next show an efficient method for identifying this edge, as well as the weight of $\second{\graph}$ without actually having to find $\second{\graph}$.

\begin{definition}[{\color{mediumblue} Blue} and {\color{crimson} red} edges]
For any graph $\graph$, tree $\tree\in\arbs{\graph}$, and edge $e=\edgeij\in\tree$, the set of \defn{{\color{mediumblue}blue} edges} $\blueedges{\graph}{e}{\tree}$ and \defn{{\color{crimson}red} edges} $\rededges{\graph}{e}{\tree}$ are defined by\footnote{We can also define
$\blueedges{\graph}{e}{\tree}$
as $\edge{i'}{j}\in\blueedges{\graph}{e}{\tree} \iff i'$ is an ancestor of $j$ in $\tree$ and 
$\rededges{\graph}{e}{\tree}$
as $\edge{i'}{j}\in\rededges{\graph}{e}{\tree} \iff i'$ is a descendant of $j$ in $\tree$.
This equivalence exists as we can only swap an incoming edge to $j$ in $\tree$ without introducing a cycle if the new edge emanates from an ancestor of $j$.
The exposition using ancestors and descendants is more similar to the exposition originally presented by \citet{CameriniFM80}.
}%

\begin{align}
    \blueedges{\graph}{e}{\tree} \defeq \{e'=&\edge{i'}{j} \mid  \weight{e'} \le \weight{e}, \nonumber\\ 
    &\tree \setminus \{e\} \cup \{e'\} \in \arbs{\graph} \}  \\
    \rededges{\graph}{e}{\tree} \defeq \{e'=&\edge{i'}{j}  \mid e'\not\in\blueedges{\graph}{e}{\tree} \}
\end{align}
An example of {\color{mediumblue} blue} and {\color{crimson} red} edges are given in \cref{fig:blue}.
\end{definition}

\begin{restatable*}[]{lemma}{lemmaOneDiff}
\label{lemma:one-diff}%
For any graph $\graph$, if $\first{\graph}=\greedy{\graph}$,
then for some $e\in\first{\graph}$ and  $e'\in\blueedges{\graph}{e}{\first{\graph}}$
\begin{equation}
    \second{\graph} = \first{\graph} \setminus \{e\} \cup \{e'\}
\end{equation}
\end{restatable*}
\begin{toappendix}
\lemmaOneDiff
\begin{proof}
By \cref{thm:second}, we have $\second{\graph}=\first{(\edgedelete{\graph}{e})}$ where $e=\edge{i}{j}$ is chosen according to \cref{eq:del-e}.
Consider the graph $\edgedelete{\graph}{e}$; we have that $\greedy{\edgedelete{\graph}{e}}=\first{\graph}\setminus\{e\}\cup\{e'\}$ where $e'$ is the second best incoming edge to $j$ in $\graph$ by the definition of the greedy graph.

\begin{enumerate}
\item \case{Case $e'\in\blueedges{\graph}{e}{\first{\graph}}$} Then $\greedy{\edgedelete{\graph}{e}}$ is a tree and $\first{(\edgedelete{\graph}{e})}=\greedy{\edgedelete{\graph}{e}}$.

\item \case{Case $e'\in\rededges{\graph}{e}{\first{\graph}}$} 
Then, $\greedy{\edgedelete{\graph}{e}}$ has a cycle $\cg$ by construction. Since this is a greedy graph, cycle $\cg$ is critical.
In the expansion phase of the $1$-best algorithm, we will break the cycle $\cg$.

\begin{enumerate}
\item \case{Case break $\cg$ at $j$} Then, $e'\not\in\first{(\edgedelete{\graph}{e})}$ and we must choose an edge $e''=\edge{i'}{j}$ to be in $\first{(\edgedelete{\graph}{e})}$. We require that $e''\in\blueedges{\graph}{e}{\first{\graph}}$ as we would otherwise re-introduce a cycle in the expansion phase, which is not possible. Therefore, $\second{\graph}=\first{\graph}\setminus\{e\}\cup\{e''\}$.

\item \case{Case break $\cg$ at $j'\neq j$} Then, there exists an edge $f=\edge{i''}{j'}\in\cg$ (and in $\first{\graph}$) which is not in $\second{\graph}$.
Instead, we choose $f'=\edge{i'}{j'}$ to be in $\second{\graph}$.
Therefore, $\second{\graph}=\first{\graph}\setminus\{e, f\}\cup\{e', f'\}$.
However, it is not possible for $f'$ and $e$ to form a cycle and so $\tree=\first{\graph}\setminus\{f\}\cup\{f'\}\in\arbs{\graph}$ and $\weight{\tree}>\weight{\second{\graph}}$.
This is a contradiction as only $\weight{\first{\graph}}>\weight{\second{\graph}}$.
\end{enumerate}

\end{enumerate}

\end{proof}
\end{toappendix}

\cref{lemma:one-diff} can be understood more clearly by following the worked example in \cref{fig:blue}.
The moral of \cref{lemma:one-diff} is that in the base case where there are no critical cycles, we only need to examine the {\color{mediumblue}blue} edges of the greedy graph to find the second best tree.
Furthermore, our second best tree will only differ from our best tree by exactly one {\color{mediumblue}blue} edge.
\citet{CameriniFM80} make use of the concepts of the {\color{mediumblue}blue} and {\color{crimson}red} edge sets, but rather than consider a base case as \cref{lemma:one-diff}, they propose an ordering in which to visit the edges of the graph.
This results in several properties about the possible orderings, requiring much more complicated proofs.

\begin{definition}[Swap cost]
For any graph $\graph$, tree $\tree\in\arbs{\graph}$, and edge $e\in\tree$, the \defn{swap cost} denotes the minimum change to a tree weight to replace $e$ by a single edge in $\tree$. It is given by
\begin{equation}
    \swapcost{\graph,\tree}{e}=\min_{e'\in\blueedges{\graph}{e}{\tree}}\left(\weight{e} - \weight{e'}\right)
\end{equation}
We will shorthand $\swapcost{\graph}{e}$ to mean $\swapcost{\graph,\first{\graph}}{e}$.
\end{definition}

\begin{cor}\label{cor:greedy-second}
For any graph $\graph$, if $\first{\graph}=\greedy{\graph}$, then $\second{\graph}=\first{(\edgedelete{\graph}{e})}$ where $e$ is given by
\begin{equation}\label{eq:min-cost}
    e=\argmin_{e'\in\first{\graph}}\swapcost{\graph}{e'}
\end{equation}
Furthermore, $\weight{\second{\graph}}=\weight{\first{\graph}} - \swapcost{\graph}{e}$.
\end{cor}

\cref{cor:greedy-second} provides us a procedure for finding the best edge to remove to find $\second{\graph}$ as well as its weight in the base case of $\graph$ having no critical cycles.
We next illustrate what must be done in the recursive case when a critical cycle exists.

\begin{figure*}[htb!]
\centering
\begin{tikzpicture}
\tikzset{cross/.style={cross out, draw=crimson, minimum size=2*(#1-\pgflinewidth), inner sep=0pt, outer sep=0pt, very thick}, cross/.default={0.5em}}

\tikzstyle{block} = [rectangle, draw, fill=white, very thick,
    minimum width=6em, rounded corners, minimum height=7em, inner sep=4pt, font=\small\linespread{1}\selectfont]

    \node [block, minimum height=6em] (g1) at (0.7, 2.7) {};
    \node (g1-t) at (0.7, 3.55) {\small $\kbest{1}{\graph}$, $w: 260$ };
    \begin{scope}[>=latex,
            every node/.style={circle,thick,draw, inner sep=1pt, font=\tiny\linespread{1}\selectfont},
            every edge/.style={draw, thick}]
        \node (r) at (0.1, 3.1) {$\root$};
        \node (1) at (0., 2.1) {$1$};
        \node (2) at (1.3, 2.9) {$2$};
        \node (3) at (1.3, 1.9) {$3$};
        \node (4) at (0.6, 2.5) {$4$};
        
        \path [->] (r) edge[\Cma] (1);
        \path [->] (r) edge[\Cma] (2);
        \path [->] (1) edge[] (3);
        \path [->] (2) edge[\Cma]  (4);
        \path [->] (2) edge[]  (3);
        \path [->] (3) edge[bend right=40] (2);
        \path [->] (4) edge[\Cma] (3);
        \path [->] (4) edge[]  (1);
    \end{scope}

    \node [block] (g2) at (3.8, 2.7) {};
    \node (g2-t) at (3.8, 3.7) {\small $\kbest{2}{\graph}$, $w: 220$ };
    \begin{scope}[>=latex,
            every node/.style={circle,thick,draw, inner sep=1pt, font=\tiny\linespread{1}\selectfont},
            every edge/.style={draw, thick}]
        \node (r) at (3.2, 3.27) {$\root$};
        \node (1) at (3.1, 2.27) {$1$};
        \node (2) at (4.4, 3.07) {$2$};
        \node (3) at (4.4, 2.07) {$3$};
        \node (4) at (3.7, 2.67) {$4$};
        
        \path [->] (r) edge[\Cma] (1);
        \path [->] (r) edge[\Cma] (2);
        \path [->] (1) edge[] (3);
        \path [->] (2) edge[\Cma]  (4);
        \path [->] (2) edge[\Cma]  (3);
        \path [->] (3) edge[bend right=40,] (2);
        \path [->] (4) edge[\Cselect] (3);
        \path [->] (4) edge[]  (1);
    \end{scope}
    \node (g2-e) at (3.8, 1.7) {\small $e: \edge{4}{3}$};
    
    \node [block] (g5) at (6.9, 4.2) {};
    \node (g5-t) at (6.9, 5.2) {\small $\kbest{5}{\graph}$, $w: 190$ };
    \begin{scope}[>=latex,
            every node/.style={circle,thick,draw, inner sep=1pt, font=\tiny\linespread{1}\selectfont},
            every edge/.style={draw, thick}]
        \node (r) at (6.3, 4.77) {$\root$};
        \node (1) at (6.2, 3.77) {$1$};
        \node (2) at (7.5, 4.57) {$2$};
        \node (3) at (7.5, 3.57) {$3$};
        \node (4) at (6.8, 4.17) {$4$};
        
        \path [->] (r) edge[\Cselect] (1);
        \path [->] (r) edge[\Cma] (2);
        \path [->] (1) edge[\Cexc] node[cross, minimum size=0.15cm] {} (3);
        \path [->] (2) edge[\Cma]  (4);
        \path [->] (2) edge[\Cexc] node[cross, minimum size=0.15cm] {} (3);
        \path [->] (3) edge[bend right=40,] (2);
        \path [->] (4) edge[\Cma] (3);
        \path [->] (4) edge[\Cma]  (1);
    \end{scope}
    \node (g5-e) at (6.9, 3.2) {\small $e: \edge{\root}{1}$};
    
    \node [block] (g3) at (6.9, 1.2) {};
    \node (g3-t) at (6.9, 2.2) {\small $\kbest{3}{\graph}$, $w: 210$ };
    \begin{scope}[>=latex,
            every node/.style={circle,thick,draw, inner sep=1pt, font=\tiny\linespread{1}\selectfont},
            every edge/.style={draw, thick}]
        \node (r) at (6.3, 1.77) {$\root$};
        \node (1) at (6.2, 0.77) {$1$};
        \node (2) at (7.5, 1.57) {$2$};
        \node (3) at (7.5, 0.57) {$3$};
        \node (4) at (6.8, 1.17) {$4$};
        
        \path [->] (r) edge[\Cma] (1);
        \path [->] (r) edge[\Cselect] (2);
        \path [->] (1) edge[\Cma] (3);
        \path [->] (2) edge[\Cma]  (4);
        \path [->] (2) edge[]  (3);
        \path [->] (3) edge[bend right=40, \Cma] (2);
        \path [->] (4) edge[\Cexc] node[cross, minimum size=0.15cm] {} (3);
        \path [->] (4) edge[]  (1);
    \end{scope}
    \node (g3-e) at (6.9, 0.2) {\small $e: \edge{\root}{2}$};
    
    \node [cross] (c1) at (9.1, 5.1) {};
    \node [cross] (c2) at (9.1, 4.1) {};
    
    \node [cross] (c3) at (9.1, 0.2) {};
    
    \node [block] (g4) at (10., 2.) {};
    \node (g4-t) at (10., 3.) {\small $\kbest{4}{\graph}$, $w: 200$ };
    \begin{scope}[>=latex,
            every node/.style={circle,thick,draw, inner sep=1pt, font=\tiny\linespread{1}\selectfont},
            every edge/.style={draw, thick}]
        \node (r) at (9.4, 2.57) {$\root$};
        \node (1) at (9.3, 1.57) {$1$};
        \node (2) at (10.6, 2.37) {$2$};
        \node (3) at (10.6, 1.37) {$3$};
        \node (4) at (9.9, 1.97) {$4$};
        
        \path [->] (r) edge[\Cma] (1);
        \path [->] (r) edge[\Cma] (2);
        \path [->] (1) edge[\Cma] (3);
        \path [->] (2) edge[\Cma]  (4);
        \path [->] (2) edge[\Cselect]  (3);
        \path [->] (3) edge[bend right=40, \Cexc] node[cross, minimum size=0.15cm] {} (2);
        \path [->] (4) edge[\Cexc] node[cross, minimum size=0.15cm] {} (3);
        \path [->] (4) edge[]  (1);
    \end{scope}
    \node (g4-e) at (10., 1.) {\small $e: \edge{2}{3}$};
    
    \node [block] (g6) at (13.1, 4.2) {};
    \node (g6-t) at (13.1, 5.2) {\small $\kbest{6}{\graph}$, $w: 150$ };
    \begin{scope}[>=latex,
            every node/.style={circle,thick,draw, inner sep=1pt, font=\tiny\linespread{1}\selectfont},
            every edge/.style={draw, thick}]
        \node (r) at (12.5, 4.77) {$\root$};
        \node (1) at (12.4, 3.77) {$1$};
        \node (2) at (13.7, 4.57) {$2$};
        \node (3) at (13.7, 3.57) {$3$};
        \node (4) at (13., 4.17) {$4$};
        
        \path [->] (r) edge[\Cselect] (1);
        \path [->] (r) edge[\Cma] (2);
        \path [->] (1) edge[\Cexc] node[cross, minimum size=0.15cm] {} (3);
        \path [->] (2) edge[\Cma]  (4);
        \path [->] (2) edge[\Cma]  (3);
        \path [->] (3) edge[bend right=40, \Cexc] node[cross, minimum size=0.15cm] {} (2);
        \path [->] (4) edge[\Cexc] node[cross, minimum size=0.15cm] {} (3);
        \path [->] (4) edge[\Cma] (1);
    \end{scope}
    \node (g6-e) at (13.1, 3.2) {\small $e: \edge{\root}{1}$};
    
    \node [block] (g7) at (13.1, 1.2) {};
    \node (g7-t) at (13.1, 2.2) {\small $\kbest{7}{\graph}$, $w: 130$ };
    \begin{scope}[>=latex,
            every node/.style={circle,thick,draw, inner sep=1pt, font=\tiny\linespread{1}\selectfont},
            every edge/.style={draw, thick}]
        \node (r) at (12.5, 1.77) {$\root$};
        \node (1) at (12.4, 0.77) {$1$};
        \node (2) at (13.7, 1.57) {$2$};
        \node (3) at (13.7, 0.57) {$3$};
        \node (4) at (13, 1.17) {$4$};
        
        \path [->] (r) edge[\Cselect] (1);
        \path [->] (r) edge[\Cma] (2);
        \path [->] (1) edge[\Cma] (3);
        \path [->] (2) edge[\Cma]  (4);
        \path [->] (2) edge[\Cexc] node[cross, minimum size=0.15cm] {} (3);
        \path [->] (3) edge[bend right=40, \Cexc] node[cross, minimum size=0.15cm] {} (2);
        \path [->] (4) edge[\Cexc] node[cross, minimum size=0.15cm] {} (3);
        \path [->] (4) edge[\Cma]  (1);
    \end{scope}
    \node (g7-e) at (13.1, 0.2) {\small $e: \edge{\root}{1}$};
    
    \node [cross] (c4) at (15.2, 5) {};
    \node [cross] (c5) at (15.2, 3.4) {};
    
    \node [cross] (c6) at (15.2, 2) {};
    \node [cross] (c7) at (15.2, 0.4) {};
    
    \begin{scope}[>=latex,
              every edge/.style={draw, very thick}]
        \path [->] (g1) edge (g2);
        \path [->] (g2) edge node[above=0.15cm] {$+ e$} (g5);
        \path [->] (g2) edge node[below=0.15cm] {$- e$} (g3);
        \path [->] (g5) edge node[above=0.15cm] {$+ e$} (c1);
        \path [->] (g5) edge node[below=0.15cm] {$- e$} (c2);
        \path [->] (g3) edge node[above=0.15cm] {$+ e$} (g4);
        \path [->] (g3) edge node[below=0.15cm] {$- e$} (c3);
        \path [->] (g4) edge node[above=0.15cm] {$+ e$} (g6);
        \path [->] (g4) edge node[below=0.15cm] {$- e$} (g7);
        \path [->] (g6) edge node[above=0.15cm] {$+ e$} (c4);
        \path [->] (g6) edge node[below=0.15cm] {$- e$} (c5);
        \path [->] (g7) edge node[above=0.15cm] {$+ e$} (c6);
        \path [->] (g7) edge node[below=0.15cm] {$- e$} (c7);
    \end{scope}
   
\end{tikzpicture}
\caption{
Example of running through $\kbestalg$ using the graph of \cref{fig:example}.
We start with $\first{\graph}$ that has a weight of $260$ and consider the best edge to remove to find $\second{\graph}$. Using $\optnext$ we find that $\second{\graph}=\first{(\edgedelete{\graph}{e})}$ for $e=\edge{4}{3}$.
We then know that either $e\in\kbest{3}{\graph}$ or $e\not\in\kbest{3}{\graph}$.
We can push these two possibilities to the queue using two calls to $\optnext$.
We find that $\kbest{3}{\graph}$ comes from the graph without $e$, and also removes the edge $e'=\edge{\root}{2}$.
We attempt to push two new elements to the queue, but we see that only by including $e'$ in the graph can we find another tree.
We repeat this process until we have found $\kbest{K}{\graph}$ or the queue is empty.
}
\label{fig:kbest}
\end{figure*}

\begin{restatable*}[]{lemma}{lemmaExpansion}
\label{lemma:expansion}%
For any $\graph$ with a critical cycle $\cg$, either $\second{\graph}\!=\!\stitch{\second{(\contract{\graph}{\cg})}}{\cg}$ (with $\weight{\second{\graph}}\!=\!\weight{\second{(\contract{\graph}{\cg})}}$) or $\second{\graph}\!=\!\first{(\edgedelete{\graph}{e})}$ (with $\weight{\second{\graph}}\!=\!\weight{\first{\graph}}-\swapcost{\graph}{e}$) for some $e\in\cg\cap\first{\graph}$.
\end{restatable*}
\begin{toappendix}
\lemmaExpansion
\begin{proof}
It must be that $\second{\graph}=\stitch{\second{(\contract{\graph}{\cg})}}{\cg}$ or $\second{\graph}\neq\stitch{\second{(\contract{\graph}{\cg})}}{\cg}$.
\begin{enumerate}
\item \case{Case $\second{\graph}=\stitch{\second{(\contract{\graph}{\cg})}}{\cg}$}
Since the weight of a tree is preserved during expansion, we are done.

\item \case{Case $\second{\graph}\neq\stitch{\second{(\contract{\graph}{\cg})}}{\cg}$}
Then, for all $e'\!\in\!\first{(\contract{\graph}{\cg})}$, $\prov(e')\!\in\!\second{\graph}$.
Therefore, if $j$ is the entrance site of $\cg$ in $\first{(\contract{\graph}{\cg})}$, $\second{\graph}\!=\!\prov(\first{(\contract{\graph}{\cg})})\cup \second{\cg_j}$.
As $\first{\cg_j}\!=\!\greedy{\cg_j}$, by \cref{cor:greedy-second},  $\second{\cg_j}\!=\!\first{(\edgedelete{\cg_j}{e})}$ for $e\!\in\!\first{\cg_j}$ and $\weight{\second{\cg_j}}\!=\!\weight{\first{\cg_j}}-\swapcost{\cg_j}{e}$.
Thus, $\second{\graph}\!=\!\first{(\edgedelete{\graph}{e})}$ where $e\!\in\!\cg\!\cap\!\tree$ and $\weight{\second{\graph}}\!=\!\weight{\first{\graph}}-\swapcost{\graph}{e}$.
\end{enumerate}
\end{proof}
\end{toappendix}

Combining \cref{cor:greedy-second} and \cref{lemma:expansion}, we can directly modify $\opt$ to find the weight of $\second{\graph}$ and the edge we must remove to obtain it.
We detail this algorithm as $\optnext$ in \cref{alg:next-best}.

\begin{restatable*}[]{thm}{thmOptnext}
\label{thm:optnext}%
For any graph $\graph$, executing $\optnextCall{\graph}$ returns $\first{\graph}$ and $\langle w, e \rangle$ such that $\second{\graph}=\first{(\edgedelete{\graph}{e})}$ and $\weight{\second{\graph}}=w$.
\end{restatable*}
\begin{toappendix}
\thmOptnext
\begin{proof}
$\optnextCall{\graph}$ returns $\first{\graph}$ by the correctness of $\opt$.
We prove that $w, e$ satisfy the above conditions.
\begin{enumerate}
\item \case{Case $\first{\graph}=\greedy{\graph}$}
Then, by \cref{cor:greedy-second} we can find the best edge to remove and the weight of $\second{\graph}$.

\item \case{Case $\first{\graph}\neq\greedy{\graph}$}
Then, $\graph$ has a critical cycle $\cg$.
By \cref{lemma:expansion},
we can either recursively call $\optnextCall{\contract{\graph}{\cg}}$ or examine the edges in $\cg\cap\first{\graph}$ to find the best edge to remove and the weight of $\second{\graph}$.
\end{enumerate}
\end{proof}
\end{toappendix}

\paragraph{Runtime analysis.}
We know that without lines \cref{line:argmin1,line:w1,line:argmin2,line:w2}, $\optnext$ is identical to $\opt$ and so will run in $\bigo{\nN^2}$.
We call $\bar{w}$ at most $\nN+2$ times during a full call of $\optnext$: $\nN$ times from lines \cref{line:argmin1,line:argmin2} combined, once from \cref{line:w1}, and once from \cref{line:w2}.
To find $\bar{w}$, we first need to find the set of {\color{mediumblue}blue} edges, which can be done in $\bigo{\nN}$ by computing the reachability graph.
Then, we need another $\bigo{\nN}$ to find the minimising value.
Therefore, $\optnext$ does $\bigo{\nN^2}$ extra work than $\opt$ and so retains the runtime of $\bigo{\nN^2}$.
\citet{CameriniFM80} require $\first{\graph}$ to be known ahead of time.
This results in having to run the original algorithm in $\bigo{\nN^2}$ time and then having to do the same amount of work as $\optnext$ because they must still contract the graph.
Therefore, $\optnext$ has a constant-time speed-up over its counterpart in \citet{camerini1979note}.\looseness=-1

\section{Finding the $\nK\th$ Best Tree}\label{sec:kbest}
In the previous section, we found an efficient method for finding $\second{\graph}$.
We now utilize this method to efficiently find the $\nK$-best trees.

\begin{restatable*}[]{lemma}{lemmaKbest}
\label{lemma:k-best}%
For any graph $\graph$ and $\nK>1$, there exists a subgraph $\graph'\subseteq\graph$ and $1 \le l < \nK$ such that $\kbest{l}{\graph}=\first{\graph'}$ and $\kbest{\nK}{\graph}=\second{\graph'}$.
\end{restatable*}
\begin{toappendix}
\lemmaKbest
\begin{proof}
There must exist some subgraph $\graph'\subseteq\graph$ such that $\kbest{\nK}{\graph}=\second{\graph'}$.
Suppose by way of contradiction that there does not exist an $l < \nK$ such that $\kbest{l}{\graph}=\first{\graph'}$.
However, since $\weight{\first{\graph'}} > \weight{\kbest{\nK+1}{\graph}}$, $\first{\graph'}$ must be in the $\nK$-highest weighted trees.
Therefore, there must exist an $l$ such that $\kbest{l}{\graph}=\first{\graph'}$
\end{proof}
\end{toappendix}

\cref{lemma:k-best} suggests that we can find the $\nK$-best trees by only examining the second best trees of subgraphs of $\graph$.
This idea is formalized as algorithm $\kbestalg$ in \cref{alg:kbest}.
A walk-through of the exploration space using $\kbestalg$ for our example graph in \cref{fig:example} is shown in \cref{fig:kbest}.

\begin{restatable*}[]{thm}{thmKbest}
For any graph $\graph$ and $\nK \!>\! 0$, at any iteration $1\le k \le K$, $\kbestalgCall{\graph}{\nK}$ returns $\kbest{k}{\graph}$.
\end{restatable*}
\begin{toappendix}
\thmKbest
\begin{proof}
We prove this by induction on $k$.

\case{Base Case}
Then, $k=1$ and $\first{\graph}$ is returned by \cref{thm:optnext}.

\case{Inductive Step}
Assume that for all $l\le k$, at iteration $l$, $\kbest{l}{\graph}$ is returned.
Now consider iteration $k+1$, by \cref{lemma:k-best}, we know that $\kbest{k+1}{\graph}=\second{\graph'}$ where $\first{\graph'}=\kbest{l}{\graph}$ for some $l \le k$.
By the induction hypothesis, $\kbest{l}{\graph}$ is returned at the $l\th$ iteration, and by \cref{thm:optnext}, we have pushed $\second{\graph'}$ onto the queue.
Therefore, we will return $\kbest{k+1}{\graph}$.
\end{proof}
\end{toappendix}

\begin{figure}[t!]
\input{figures/kbest_alg}
\caption{$\nK$-best tree enumeration algorithm.}
\label{alg:kbest}
\end{figure}

\begin{figure*}[htb!]
    \centering
    \tikzset{cross/.style={cross out, draw=crimson, minimum size=2*(#1-\pgflinewidth), inner sep=0pt, outer sep=0pt, very thick},
cross/.default={0.5em}}

\tikzstyle{block} = [rectangle, draw, fill=white, very thick,
    minimum width=6em, rounded corners, minimum height=7em, inner sep=4pt, font=\small\linespread{1}\selectfont]

\begin{tikzpicture}

    \node [block] (g1) at (0.7, 2.7) {};
    \node (g1-t) at (0.7, 3.7) {\small $\kbestrc{1}{\graph}$, $w: 210$ };
    \begin{scope}[>=latex,
            every node/.style={circle,thick,draw, inner sep=1pt, font=\tiny\linespread{1}\selectfont},
            every edge/.style={draw, thick}]
        \node (r) at (0.1, 3.27) {$\root$};
        \node (1) at (0., 2.27) {$1$};
        \node (2) at (1.3, 3.07) {$2$};
        \node (3) at (1.3, 2.07) {$3$};
        \node (4) at (0.6, 2.67) {$4$};
        
        \path [->] (r) edge[\Cma] (1);
        \path [->] (r) edge[] (2);
        \path [->] (1) edge[\Cma] (3);
        \path [->] (2) edge[\Cma]  (4);
        \path [->] (2) edge[]  (3);
        \path [->] (3) edge[bend right=40, \Cma] (2);
        \path [->] (4) edge[] (3);
        \path [->] (4) edge[]  (1);
    \end{scope}
    \node (g1-e) at (0.7, 1.7) {\small $e: \edge{\root}{1}$};
    
    \node [cross] (c1) at (3.2, 3.5) {};
    
    \node [block] (g2) at (4.2, 1.5) {};
    \node (g2-t) at (4.2, 2.5) {\small $\kbestrc{2}{\graph}$, $w: 190$ };
    \begin{scope}[>=latex,
            every node/.style={circle,thick,draw, inner sep=1pt, font=\tiny\linespread{1}\selectfont},
            every edge/.style={draw, thick}]
        \node (r) at (3.6, 2.07) {$\root$};
        \node (1) at (3.5, 1.07) {$1$};
        \node (2) at (4.8, 1.87) {$2$};
        \node (3) at (4.8, 0.87) {$3$};
        \node (4) at (4.1, 1.57) {$4$};
        
        \path [->] (r) edge[\Cexc] node[cross, minimum size=0.15cm] {} (1);
        \path [->] (r) edge[\Cma] (2);
        \path [->] (1) edge[] (3);
        \path [->] (2) edge[\Cma]  (4);
        \path [->] (2) edge[]  (3);
        \path [->] (3) edge[bend right=40] (2);
        \path [->] (4) edge[\Cma] (3);
        \path [->] (4) edge[\Cma]  (1);
    \end{scope}
    \node (g2-e) at (4.2, 0.5) {\small $e: \edge{\root}{2}$};
    
    \node [cross] (c2) at (6.7, 0.7) {};
    
    \node [block] (g3) at (7.7, 2.7) {};
    \node (g3-t) at (7.7, 3.7) {\small $\kbestrc{3}{\graph}$, $w: 150$ };
    \begin{scope}[>=latex,
            every node/.style={circle,thick,draw, inner sep=1pt, font=\tiny\linespread{1}\selectfont},
            every edge/.style={draw, thick}]
        \node (r) at (7.1, 3.27) {$\root$};
        \node (1) at (7., 2.27) {$1$};
        \node (2) at (8.3, 3.07) {$2$};
        \node (3) at (8.3, 2.07) {$3$};
        \node (4) at (7.6, 2.67) {$4$};
        
        \path [->] (r) edge[\Cexc] node[cross, minimum size=0.15cm] {} (1);
        \path [->] (r) edge[\Cma] (2);
        \path [->] (1) edge[] (3);
        \path [->] (2) edge[\Cma]  (4);
        \path [->] (2) edge[\Cma]  (3);
        \path [->] (3) edge[bend right=40, \Cexc] node[cross, minimum size=0.15cm] {}  (2);
        \path [->] (4) edge[\Cselect] (3);
        \path [->] (4) edge[\Cma]  (1);
    \end{scope}
    \node (g3-e) at (7.7, 1.7) {\small $e: \edge{4}{3}$};
    
    \node [cross] (c3) at (10.2, 3.5) {};
    
    \node [block] (g4) at (11.2, 1.5) {};
    \node (g4-t) at (11.2, 2.5) {\small $\kbestrc{4}{\graph}$, $w: 130$ };
    \begin{scope}[>=latex,
            every node/.style={circle,thick,draw, inner sep=1pt, font=\tiny\linespread{1}\selectfont},
            every edge/.style={draw, thick}]
        \node (r) at (10.6, 2.07) {$\root$};
        \node (1) at (10.5, 1.07) {$1$};
        \node (2) at (11.8, 1.87) {$2$};
        \node (3) at (11.8, 0.87) {$3$};
        \node (4) at (11.1, 1.57) {$4$};
        
        \path [->] (r) edge[\Cexc] node[cross, minimum size=0.15cm] {} (1);
        \path [->] (r) edge[\Cma] (2);
        \path [->] (1) edge[\Cma] (3);
        \path [->] (2) edge[\Cma]  (4);
        \path [->] (2) edge[\Cselect]  (3);
        \path [->] (3) edge[bend right=40, \Cexc] node[cross, minimum size=0.15cm] {} (2);
        \path [->] (4) edge[\Cexc] node[cross, minimum size=0.15cm] {} (3);
        \path [->] (4) edge[\Cma]  (1);
    \end{scope}
    \node (g4-e) at (11.2, 0.5) {\small $e: \edge{2}{3}$};
    
    \node [cross] (c4) at (13.7, 2.2) {};
    \node [cross] (c5) at (13.7, 0.7) {};

    \begin{scope}[>=latex,
              every edge/.style={draw, very thick}]
        \path [->] (g1) edge node[above] {$+ e$} (c1);
        \path [->] (g1) edge node[below=0.15cm] {$- e$} (g2);
        \path [->] (g2) edge node[above] {$+ e$} (g3);
        \path [->] (g2) edge node[below=0.15cm] {$- e$} (c2);
        \path [->] (g3) edge node[above] {$+ e$} (c3);
        \path [->] (g3) edge node[below=0.15cm] {$- e$} (g4);
        \path [->] (g4) edge node[above] {$+ e$} (c4);
        \path [->] (g4) edge node[below=0.15cm] {$- e$} (c5);

    \end{scope}
    
\end{tikzpicture}
    \caption{
Example of running through $\kbestrcalg$ using the graph of \cref{fig:example}.
We start with $\kbestrc{1}{\graph}$ that has a weight of $210$ and consider the best edge to remove to find $\second{\graph}$.
We consider removing the best dependency tree with the same edge emanating from the root $e=\edge{\root}{1}$ using $\optnext$.
However, no such dependency tree exists, and so we only need to push the graph $\edgedelete{\graph}{e}$.
When we next pop from the queue, we see that we have removed root edge $e$, and so must consider removing the new root edge $e'=\edge{\root}{e}$.
In this case, no dependency tree exists without $e$ and $e'$, and so we only push to the queue the results of running $\optnext$.
We repeat this process until we have found $\kbestrc{\nK}{\graph}$ or the queue is empty.
}
    \label{fig:kbest-rc}
\end{figure*}

\begin{table}[t!]
    \centering
    \begin{tabular}{lcccc}
         & $\nK=10$ & $\nK=20$ & $\nK=50$ \\ \midrule
         \citeauthor{CameriniFM80} & $6.95\hphantom{\times}$ & $14.04\hphantom{\times}$ & $35.11\hphantom{\times}$ \\
         $\kbestalg$ & $4.89\hphantom{\times}$ & $10.10\hphantom{\times}$ & $25.63\hphantom{\times}$ \\ \midrule
         Speed-up & $1.42\times$ & $\hphantom{1}1.39\times$ & $\hphantom{1}1.37\times$ 
    \end{tabular}
    \caption{Runtime experiment for parsing the $K$-best spanning trees in the English UD test set \citep{ud}.
    Times are given in $10^{-2}$ seconds for the average parse of the $K$-best spanning trees.
    }
    \label{tab:runtime}
\end{table}

\noindent\paragraph{Runtime analysis.}
We call $\optnext$ once at the start of the algorithm, then every subsequent iteration we make two calls to $\optnext$.
As we have $K-1$ iterations
, the runtime of $\kbestalg$ is $\bigo{\nK \nN^2}$.
The first call to $\optnext$ in each iteration finds the $K\th$ best tree as well as an edge to remove.
\citet{CameriniFM80} make one call to of $\opt$ and two calls to $\optnext$ which only finds the weight-edge pair of our algorithm.
Therefore, $\kbestalg$ has a constant time speed-up on the original algorithm.\footnote{In practice, we maintain a set of edges to include and exclude to save space.
}

\noindent\paragraph{A short experiment.}
We empirically measure the constant time speed-up between $\kbestalg$ and the original algorithm of \citet{CameriniFM80}.
We take the English UD test set (as used for \cref{fig:experiment}) and find the $10$, $20$, and $50$ best spanning trees using both algorithms.\footnote{Implementations for both versions can be found in our code release (see \cref{footnote:code})}
We give the results of the experiment in \cref{tab:runtime}.\footnote{The experiment was conducted using an Intel(R) Core(TM) i7-7500U processor with 16GB RAM.}
We note that on average $\kbestalg$ leads to a $1.39$ times speed-up.
This is lower than we anticipated as we have to make half as many calls to $\optnext$ than the original algorithm.
However, in the original $\optnext$ of \citet{CameriniFM80}, we do not require to stitch together the tree, which may explain the slightly smaller speed-up.

\section{Finding the $\nK\th$ Best \emph{Dependency} Tree}\label{sec:kbest-rc}
In this section, we present a novel extension to the algorithm presented thus far, that allows us to efficiently find the $\nK$-best dependency trees.
Recall that we consider dependency trees to be spanning trees \emph{with a root constraint} such that only one edge may emanate from $\root$.
\Naively, we can use $\kbestalg$ where we initialize the queue with $\first{(\edgeinclude{\graph}{e_{\root}})}$ for each $e_{\root}=\edge{\root}{j}\in\graph$.
However, this adds a $\bigo{\nN^3}$ component to our runtime as we have to call $\opt$ $\nN$ times.
Instead, our algorithm maintains the $\bigo{\nK \nN^2}$ runtime as the regular $\nK$-best algorithm.
We begin by noting that we can find second best dependency tree, by finding either the best dependency tree with a different root edge or the second best tree with the same root edge.

\begin{restatable*}[]{lemma}{lemmaSecondRc}
\label{lemma:second-rc}%
For any graph $\graph$ and edge $e_{\root}=\edge{\root}{j}\in\kbestrc{1}{\graph}$, $\kbestrc{2}{\graph}=\kbestrc{1}{(\edgedelete{\graph}{e_{\root}})}$ or $\kbestrc{2}{\graph}=\kbestrc{2}{(\edgeinclude{\graph}{e_{\root}})}$.
\end{restatable*}
\begin{toappendix}
\lemmaSecondRc
\begin{proof}
If $e_{\root}\!\not\in\!\kbestrc{2}{\graph}$, then clearly $\kbestrc{2}{\graph}\!=\!\kbestrc{1}{(\edgedelete{\graph}{e_{\root}})}$.
Otherwise, $e_{\root}\in\kbestrc{2}{\graph}$. As $e_{\root}\in\kbestrc{1}{\graph}$, $\kbestrc{2}{\graph}=\kbestrc{2}{(\edgeinclude{\graph}{e_{\root}})}$.
\end{proof}
\end{toappendix}

\begin{restatable*}[]{lemma}{lemmaKbestRc}
\label{lemma:k-best-rc}%
For any graph $\graph$ and $\nK>1$, if $e=\edge{\root}{j}\in\kbestrc{\nK}{\graph}$, then either $e$ is not in any of the $\nK\!-\!1$-best trees or there exists a subgraph $\graph'\subseteq\graph$ and $1 \le l < \nK$ such that $\kbestrc{l}{\graph}=\kbestrc{1}{\graph'}$, $e\in\kbestrc{1}{\graph'}$ and $\kbestrc{\nK}{\graph}=\kbestrc{2}{\graph'}$.
\end{restatable*}
\begin{toappendix}
\lemmaKbestRc
\begin{proof}
It must be that either there exists an $1\le l < \nK$ such that $e\in\kbestrc{l}{\graph}$ (Case 1) or no such $l$ exists (Case 2).

\begin{enumerate}
\item Consider the graph $\edgeinclude{\graph}{e}$.
Under our definition of edge-inclusion graphs for dependency trees, $\arbs{\edgeinclude{\graph}{e}}=\arbsrc{\edgeinclude{\graph}{e}}$.
Then, by \cref{lemma:k-best}, there exists a $l'$ and $\graph'$ such that $\kbestrc{l'}{(\graph}=\kbestrc{1}{\graph'}$ and $\kbestrc{\nK}{\graph}=\kbestrc{2}{\graph'}$.

\item Then, $e$ is not in any of the $(\nK\!-\!1)$-best trees.
\end{enumerate}
\end{proof}
\end{toappendix}

\cref{lemma:k-best-rc} suggests that we can find the $\nK$-best dependency trees, by examining the second best dependency trees of subgraphs of $\graph$ or finding the best dependency tree with a unique root edge.
This idea is formalized as algorithm $\kbestrcalg$ in \cref{alg:kbest-rc}.
A walk-through of the exploration space using $\kbestrcalg$ for our example graph in \cref{fig:example} is shown in \cref{fig:kbest-rc}.

\begin{figure}[t!]
\input{figures/kbest_rc_alg}
\caption{$\nK$-best dependency tree enumeration algorithm.}
\label{alg:kbest-rc}
\end{figure}

\begin{restatable*}[]{thm}{thmKbestRc}
For any graph $\graph$ and $\nK \ge 1$, at iteration $1\le k \le K$, $\kbestrcalgCall{\graph}{\nK}$ returns $\kbestrc{k}{\graph}$.
\end{restatable*}
\begin{toappendix}
\thmKbestRc
\begin{proof}
We prove this by induction on $k$.

\case{Base Case}
Then, $k=1$ and $\first{\graph}$ is returned by the correctness of $\opt$.

\case{Inductive Step}
Assume that for all $l\le k$, at iteration $l$, $\kbestrc{l}{\graph}$ was returned.
Now consider iteration $k+1$, by \cref{lemma:k-best-rc}, we know that either $\kbestrc{k+1}{\graph}$ has a unique root edge to the $k$-best trees (Case 1) or $e=\edge{\root}{j}$ and there exists a $\graph'$ and $l \le k$ such that $\first{\graph'}=\kbest{l}{\graph}$, $e\in\kbest{l}{\graph}$, and $\kbestrc{k+1}{\graph}=\kbestrc{2}{\graph'}$ (Case 2).

\begin{enumerate}
\item There always exists a tree in the queue that has a unique root edge to all trees that came before it.
Furthermore, it is the highest such tree by the correctness of $\opt$.

\item
By our induction hypothesis, $\kbestrc{l}{\graph}$ is returned at the $l\th$ iteration, and by \cref{thm:optnext}, we have pushed $\kbestrc{2}{\edgeinclude{\graph'}{e}}$ onto the queue.
Therefore, we will return $\kbestrc{k+1}{\graph}$.
\end{enumerate}
\end{proof}
\end{toappendix}

\noindent\paragraph{Runtime analysis.}
At the start of the algorithm, we call $\opt$ twice and $\optnext$ once.
Then, at each iteration we either make two calls two $\optnext$, or two calls to $\opt$ and one call to $\optnext$.
As both algorithms have a runtime of $\bigo{\nN^2}$, each iteration has a runtime of $\bigo{\nN^2}$.
Therefore, running $\nK$ iterations gives a runtime of $\bigo{\nK \nN^2}$.

\section{Conclusion}
In this paper, we provided a simplification to \citet{CameriniFM80}'s $\bigo{\nK \nN^2}$ $\nK$-best spanning trees algorithm.
Furthermore, we provided a novel extension to the algorithm that decodes the $\nK$-best dependency trees in $\bigo{\nK \nN^2}$.
We motivated the need for this new algorithm as using regular $\nK$-best decoding yields up to $36\%$ trees which violation the root constraint.
This is a substantial (up to $44$ times) increase in the violation rate from decoding the one-best tree, and thus such an algorithm is even more important than in the one-best case.
We hope that this paper encourages future research in $\nK$-best dependency parsing.

\section*{Acknowledgments}
We would like to thank the reviewers for their valuable feedback and suggestions to improve this work. The first author is supported by the University of Cambridge School of Technology Vice-Chancellor's Scholarship as well as by the University of Cambridge Department of Computer Science and Technology's EPSRC.

\section*{Ethical Concerns}
We do not foresee how the more efficient algorithms presented this work exacerbate any existing ethical concerns with NLP systems.

\bibliography{acl2021}
\bibliographystyle{acl_natbib}

\end{document}